\pdfoutput=1

\documentclass[11pt]{article}

\usepackage{emnlp2021}

\usepackage{times}
\usepackage{latexsym}

\usepackage[T1]{fontenc}

\usepackage[utf8]{inputenc}

\usepackage{microtype}

\usepackage{amsmath}
\usepackage{graphicx}

\usepackage{booktabs}
\usepackage{multirow}
\usepackage{tabulary}

\title{Sentence Bottleneck Autoencoders from Transformer Language Models}

\author{
Ivan Montero$^{\clubsuit}$ \quad
Nikolaos Pappas$^{\clubsuit}$ \quad
Noah A. Smith$^{\clubsuit}$$^{\diamondsuit}$
\vspace{5mm}
\\
$^{\clubsuit}$Paul G. Allen School of Computer Science \& Engineering, University of Washington \\
   $^{\diamondsuit}$Allen Institute for Artificial Intelligence \\
   {\tt \{ivamon,npappas,nasmith\}@cs.washington.edu } 
}

\begin{document}
\maketitle

\begin{abstract} %

Representation learning for text via pretraining a language model on a large corpus has become a standard starting point for building NLP systems.  This approach stands in contrast to autoencoders, also trained on raw text, but with the objective of learning to encode each input as a vector that allows full reconstruction.  Autoencoders are attractive because of their latent space structure and generative properties. We therefore explore the construction of a sentence-level autoencoder from a pretrained, frozen transformer language model. We adapt the masked language modeling objective as a generative, denoising one, while only training a sentence bottleneck and a single-layer modified transformer decoder.
We demonstrate that the sentence representations discovered by our model achieve better quality than previous methods that extract representations from pretrained transformers on text similarity tasks, style transfer (an example of controlled generation), and single-sentence classification tasks in the GLUE benchmark, while using fewer parameters than large pretrained models.\footnote{Our code is available at: \url{https://github.com/ivanmontero/autobot}}

\end{abstract}

\section{Introduction} 


Recent research has focused on devising new unsupervised pretraining methods from unlabeled data that involves some form of language modeling, primarily autoregressive \citep{peters-etal-2018-deep,radford2019language}, masked \citep{devlin-etal-2019-bert, liu2019RoBERTa, conneau2020unsupervised} and generalized \citep{radford2019language, brown2020language,pmlr-v97-song19d}, with much success on downstream tasks.  Under the hood, most of these methods use transformers \citep{vaswani17} for encoding text sequences, which allows them to learn powerful contextual word representations that have been used widely for building models in NLP. However, this  does not hold for sentence representations derived from pretrained transformer language models based on a special token or basic pooling operations.
To this end, representation learning methods have been designed to better capture semantic information from pretrained transformer language models, e.g., using Siamese networks trained with a triplet loss \cite{Reimers2019SentenceBERT} or transforming the desired sentence distribution to a Gaussian distribution through normalizing flows \cite{li-etal-2020-sentence}. 

Existing sentence representations directly derived from pretrained language models or learned by specialized methods cannot guarantee perfect reconstruction of the input, a property that can enhance the structure of their semantic space and enable their use for controlled generation tasks. For the latter, a few recent studies have looked into ways to steer generation of pretrained language models towards a particular style \citep{Dathathri2020Plug,krause2021gedi}, although they require following the gradient during the sampling process and rely on style text classifiers which might not be always available. 
The latent space of a text autoencoder allows one to perform controlled text generation by directly manipulating  sentence representations using basic numerical operations \cite{pmlr-v119-shen20c}. Yet, how to convert pretrained transformer language models to autoencoders with such properties still remains unexplored. 

\begin{figure}[ht]
\centering
\hspace{-2mm}\includegraphics[width=0.4\textwidth]{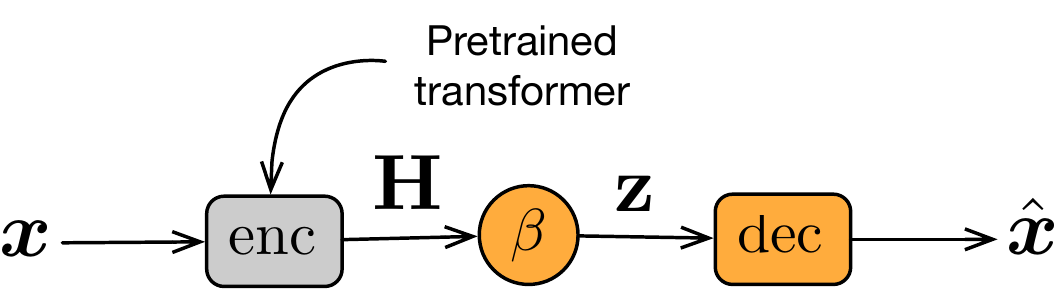} 
\caption{\label{fig:model} 
Our autoencoder consists of a pretrained transformer encoder $\text{enc}$, a function $\beta$ that compresses the encoder's  final representations $\mathbf{H}$ of size $T \times d$ 
to a sentence bottleneck representation $\mathbf{z}$ of size $d$, and a transformer decoder $\text{dec}$ that is trained to fully reconstruct the training sentence $\boldsymbol{x}$. }
\end{figure}

To fill in this gap, we introduce \textsc{Autobot}, a new autoencoder model for learning sentence
``bottleneck'' (i.e., fixed-size) representations from pretrained transformers that is useful for similarity,  generation, and classification, displayed in Figure~\ref{fig:model}. Our model has two unique components: (i) a  transformation that uses dot product attention to dynamically pool semantic information from the pretrained model's hidden states into a sentence bottleneck representation, and (ii)  a shallow transformer decoder that is modified to operate based on the bottleneck representation. Instead of training our autoencoder from scratch, we directly finetune it using an input reconstruction objective on the unlabeled data on which the original pretrained transformer was trained. We keep the underlying pretrained transformer encoder fixed, which makes it more efficient than training from scratch
and proves beneficial even if we compare to pretrained transformers trained for an equal number of steps. 

Our evaluation on representative sentence similarity, classification, and generation tasks demonstrates that the   resulting sentence representations are compact, better capture semantic similarity at the sentence-level than strong sentence representation methods \cite{Reimers2019SentenceBERT}, and can be used for controlled generation tasks.  Lastly, our model performs almost on par with the large RoBERTa model \citep{liu2019RoBERTa} even though it only introduces 1.6$\%$ additional
parameters relative to the base RoBERTa model.

\section{Model: \textsc{Autobot}}\label{sec:model}
Taking inspiration from recent research on text autoencoders \citep{bowman2015generating,shen2019educating,mai2020plug}, we extend standard autoregressive text autoencoders, which have been predominantly based on recurrent networks, to a transformer-based architecture and integrate them with pretrained language models; here we focus on RoBERTa \cite{liu2019RoBERTa}.

Autoencoders generally follow the encoder-decoder model structure to reconstruct their input with the constraint that the encoder produces a single, fixed-length hidden representation $\text{enc}(\boldsymbol{x}) = \mathbf{z}$:
\begin{equation}
\text{AE}(\boldsymbol{x}) = \text{dec}(\text{enc}(\boldsymbol{x})) = \boldsymbol{x}'.
\end{equation}
\noindent Here, we focus on denoising autoencoders that aim to reconstruct a perturbed version of the input  \citep{vincent2010stacked,shen2019educating}, which is compatible with many of the pretrained language models that are based on masked language modeling. In our experiments, we use the same masking procedure as \citet{devlin-etal-2019-bert} to perturb the input.
\subsection{Encoder}
Standard approaches use encoders that reduce the input to a single representation $\mathbf{z}$. To use a pretrained transformer for this purpose we need to reduce its output hidden representations $\mathbf{H}$ after processing the input to a single vector. Since using the special token representation or basic pooling methods have been shown sub-optimal in prior work \cite{Reimers2019SentenceBERT}, here we opt to keep the original encoder fixed and train a transformation $\beta$ that will learn to compress $\mathbf{H}$ into a single representation   $\mathbf{z} = \beta(\mathbf{H}; \mathbf{\theta})$, with $\theta$ being an additional set of parameters to be learned during finetuning. We choose $\beta$ to be a multi-head attention mechanism that takes as input the keys  $\mathbf{K}$ and values $\mathbf{V}$ corresponding to the final representations $\mathbf{H}$ from the pretrained model
and a query vector $\mathbf{q}$ corresponding to a context vector $\mathbf{u}$ that we choose to be the \texttt{CLS} vector from the pretrained model:
\begin{align}
   \beta(\mathbf{H}; \mathbf{\theta}) &= \text{MultiHead}(\mathbf{q}, \mathbf{K}, \mathbf{V}) 
\end{align}
\noindent where the parameters to be learned, $\theta$, include the weights that are used to transform the query, keys, and values which amount to $ 3 d^2 $ with $d$ being the dimensionality of each head ($d = 64$ in our experiments).

\subsection{Decoder}

The 
cross-attention layer in the Transformer decoder architecture by \citet{vaswani17}  expects hidden representations for every token input from the encoder in order for each output candidate to attend to each input token. In the situation where only a single representation comes from the encoder, we have
\begin{equation}
    \text{Attention}(\mathbf{Q}, \mathbf{z}^\top \mathbf{W}_K, \mathbf{z}^\top \mathbf{W}_V) = \mathbf{z}^\top \mathbf{W}_V
\end{equation}
Note that the queries $\mathbf{Q}$, which come from the previous masked self-attention layer, are not  taken into account, and each step in the decoder will receive the exact same $\mathbf{z}^\top \mathbf{W}_V$ as a result.  In order to mitigate this, we propose a gating method inspired by \citet{hochreiter1997long}. 
Concretely, let $\mathbf{Q_t}$ be the $t$th query representation. Then, the $t$th output $\mathbf{o}_t$ of the 
cross-attention layer is computed as follows 
\begin{equation}
    \mathbf{g}_t = \sigma(\mathbf{G}  \mathbf{Q}_t + \mathbf{G'}  \mathbf{z});\ \  
        \mathbf{o_t} = \mathbf{g}_t \odot \mathbf{z}^\top\mathbf{W}_V
\end{equation}
\noindent where $\sigma(\cdot)$ is the \textit{sigmoid} activation function and $\mathbf{G}$ and $\mathbf{G}'$
are the parameters of the transformation for the gate. 
One can view the role of the gate as determining the amount of per-element information from the linear transformation of the latent representation to keep for the current layer and timestep. Preliminary experiments found this method beneficial for generation.

\paragraph{Training considerations} To avoid training our model from scratch, we finetune it for 100K optimization  steps on a pretraining dataset using the base RoBERTa model \cite{liu2019RoBERTa} on the encoder side and a single layer 
decoder side for efficiency purposes \citep{kasai2021deep}. The model is trained using an input reconstruction loss by minimizing the negative log-likelihood computed over the reconstructed inputs. Note that only the parameters of the sentence bottleneck and the decoder are learned; the encoder parameters are kept fixed.



\section{Experiments}\label{sec:experiments}
To assess the quality of the sentence representations learned by our model we evaluate on sentence similarity (Section \ref{sec:sim}), classification (Section \ref{sec:clf}), and generation tasks (Section \ref{sec:gen}).


\subsection{Settings} 
\paragraph{Datasets}
Since the RoBERTa dataset is not publicly available, we use for pretraining the exact same dataset as BERT \citep{devlin-etal-2019-bert}, which is composed of BooksCorpus \citep{zhu2015aligning} and English Wikipedia. For sentence similarity, we use the Natural Language Inference (NLI) dataset \citep{bowman2015snli} for finetuning and evaluate on the Semantic Textual Similarity (STS) dataset \citep{cer2017semeval}, following \citet{conneau2017supervised}.  For classification, we use mainly single-sentence datasets from the GLUE benchmark \citep{wang2018glue}, namely Stanford Sentiment Treebank (SST) and Corpus of Linguistic Acceptability (CoLA) datasets, but we also report the average performance on the remaining datasets. For generation, we use the Yelp reviews dataset \citep{shen2017style}.


\paragraph{Baselines}
For sentence similarity, we compare to SBERT which is a competitive method for deriving informative sentence representations from pretrained language models \cite{Reimers2019SentenceBERT}.
They obtain sentence representations by using simple pooling methods over BERT representations such as mean and max (instead of the CLS token representation) then finetuning the whole pretrained model using Siamese networks on a combination of natural language inference data.  To compare with them on sentence similarity, we incorporate our model within their framework and follow their settings and training/evaluation protocol (details in Appendix~\ref{adx:sentrep}). 

For sentence classification, we compare our model to RoBERTa-base and RoBERTa-large models \cite{liu2019RoBERTa}. Note that BART \cite{lewis2019bart} achieves similar results to RoBERTa, so a similar comparison can be made.  

For sentence generation tasks, we compare to a strong and efficient style transfer method by \citet{shen2019educating}, which is a recurrent network-based denoising text autoencoder on in domain data. The style transfer is achieved  through vector arithmetic, namely computing a “sentiment vector” $\mathbf{v}$ by taking the vector difference between 100 negative and positive sentences, then evaluating by taking an input sentence, encoding it, adding a multiple of the sentiment vector to the sentence representation, then decoding the resulting representation. In addition to the denoising auto encoder (DAE) of \citet{shen2019educating}, we include more sophisticated methods for style transfer that are more computationally expensive such as fast gradient iterative modification (FGIM) of \citet{wang2019controllable} and Emb2Emb of \citet{mai2020plug} for reference.

\subsection{Sentence Similarity} \label{sec:sim}

 The results on the sentence similarity task are displayed in Table~\ref{tab:nli_sts}.
 Due to resource constraints and unreported results by prior work, we report our model only with RoBERTa-base.
 We can observe that \textsc{Autobot} applied to RoBERTa-base significantly outperforms other supervised base transformer methods. Additionally,  \textsc{Autobot} approaches the performance of large  transformers while having a minimal   parameter   overhead of 1.6\%.    

We also find that \textsc{Autobot} without any supervision (\textsc{Autobot}-base unsup.) outperforms all of the unsupervised methods, and most notably improves upon average BERT embeddings by 26.1\%. This demonstrates that our approach is effective in both supervised and unsupervised settings.
 

\begin{table}[t]
	\centering 
	\footnotesize
	\renewcommand{\arraystretch}{1.3}
	\begin{tabular}{l | c | c}
		\toprule
		\textbf{Model} & \textbf{Spearman} & \textbf{Parameters} \\ \midrule
		\multicolumn{3}{l}{\textit{Unsupervised}} \\\midrule
		Avg.\ GloVe embeddings & 58.02 & - \\
		Avg.\ BERT embeddings &  46.35 & - \\
		\textsc{Autobot}-base unsup. & \textbf{58.49} & - \\\midrule
		\multicolumn{3}{l}{\textit{Supervised}} \\\midrule
		InferSent - GloVe &  68.03 & - \\
		Universal Sentence Encoder &  74.92 & - \\

        RoBERTa-base & 75.37 & 125M\\
		SRoBERTa-base & 76.89 & 125M \\
		\textsc{Autobot}-base (ours) & \textbf{78.59} & 127M \\\hline
        RoBERTa-large & 80.16 & 355M \\
		 
		
		
		
		
    \bottomrule
	\end{tabular}
	\caption{ \label{tab:nli_sts}On semantic textual similarity (STS), \textsc{Autobot} outperforms previous sentence representation methods and reaches a score similar to RoBERTa-large while having fewer parameters.   
	We report Spearman's rank correlation on the test set and the model sizes are reported in terms of trained parameter size.}

	
	
\end{table}

   \begin{table}[t]
	\centering 
	\footnotesize
	\begin{tabular}{l|c}
		\toprule
		\textbf{Pooling} & \textbf{Spearman} \\ \midrule
		\texttt{MEAN} & 80.78 \\
		\texttt{MAX} & 78.76 \\
		\texttt{CLS} &  79.67 \\
		$\beta$ \text{ (ours)} & \textbf{81.88} \\
		\bottomrule
	\end{tabular}
	\caption{Performance of sentence representations from RoBERTa trained with different pooling methods on NLI data and then evaluated on STS benchmark's development set 
	in terms of Spearman's rank correlation.}
	\label{tab:pooling}
\end{table}

%

 
We find in Table~\ref{tab:pooling} that using the proposed sentence bottleneck based on learned context  provides 
noticeable gains over using simpler pooling methods from prior work. We suspect this is due to the additional flexibility provided by our bottleneck acting as ``weighted pooling'' by attending over all tokens to compute the final representation, as opposed to equal contribution of all tokens regardless of the input.



\subsection{Sentence Classification} \label{sec:clf}

The results on single-sentence classification tasks and other tasks from the GLUE benchmark are displayed in Table \ref{tab:glue}. We find that \textsc{Autobot} provides a 
noticable performance increase on single-sentence tasks, specifically on the CoLA datasets when using both the RoBERTa-base and RoBERTa-large models.
Additionally, we also find that \textsc{Autobot}, when fed both sentences concatenated for dual sentence GLUE tasks, maintains the original performance of the underlying pretrained encoder. 
\begin{table}[t]
\centering
\footnotesize
\renewcommand{\arraystretch}{1.3}
\begin{tabular}{l | c | c | c}
\toprule
\textbf{Model} & \bf SST & \bf CoLA & \bf Others (avg) \\
\midrule 
RoBERTa-base & 94.8 & 63.6 & 88.7 \\
\textsc{Autobot}-base & \textbf{95.0} & \textbf{66.0} & 88.7 \\
\midrule 
RoBERTa-large & 96.4 & 68.0 & 91.1 \\
\textsc{Autobot}-large & \textbf{96.9} & \textbf{70.2} & 91.1 \\
\bottomrule
\end{tabular} 
\caption{
Single-sentence GLUE classification dev.~results. Median accuracy is reported over over three random seeds. Our model improves performance on single-sentence classification tasks over both base and large RoBERTa models while maintaining their performance on the remaining multi-sentence tasks. 
}

\label{tab:glue}
\end{table}

 Hence, our model improves the quality of the sentence representations from pretrained transformer models without hurting their performance.

\subsection{Sentence Generation} \label{sec:gen}



\begin{figure}[ht]
\centering
\hspace{-2mm}\includegraphics[width=0.49\textwidth]{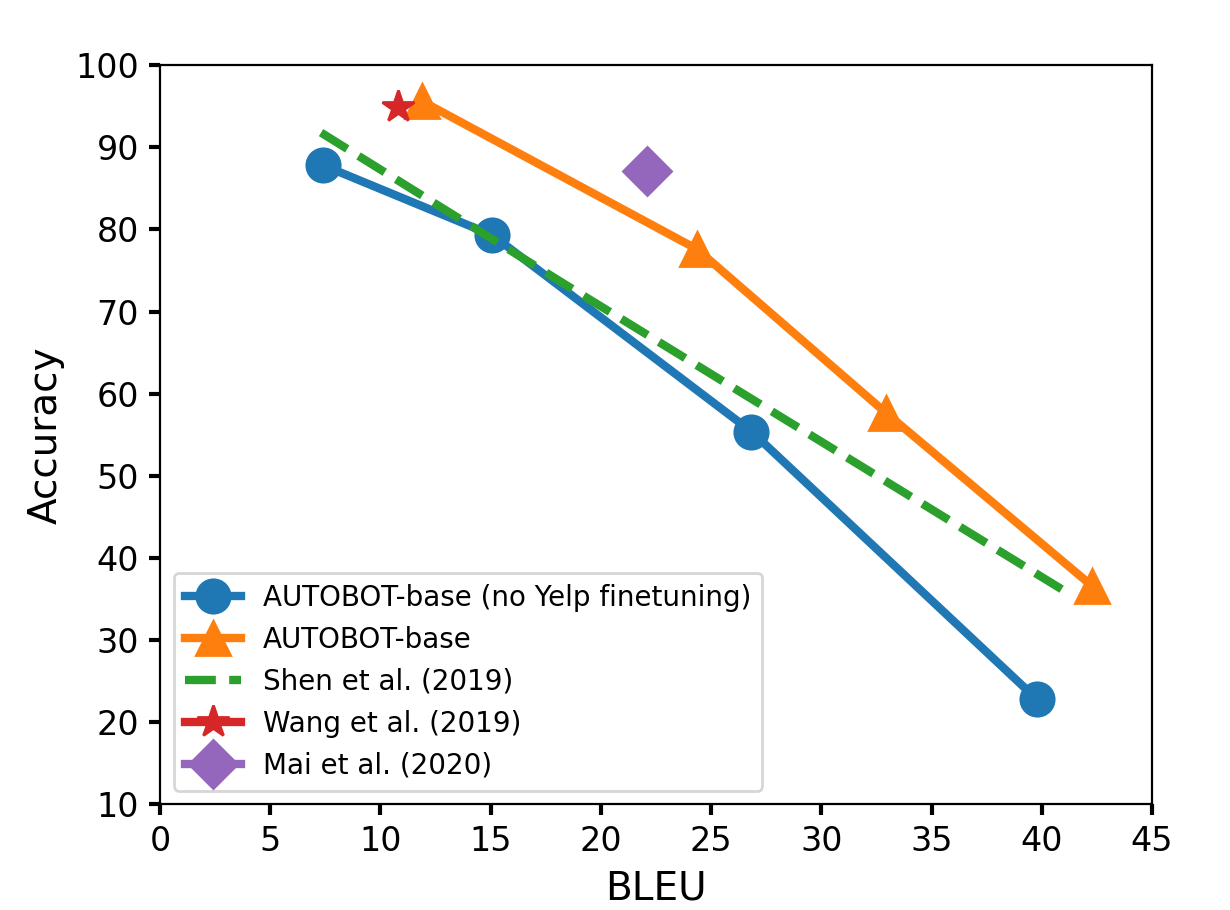}
\caption{ 
Automatic evaluations of vector arithmetic for sentiment transfer, plotted as accuracy vs. self-BLEU. Accuracy (ACC) is measured by a sentiment classifier, and values for varying multiples of the sentiment vector are plotted. Upper right is better.
\label{fig:generation}
}
\end{figure}

For sentence generation, we focus on the sentiment transfer task proposed by \citet{shen2019educating} both with and without further training on in-domain data from Yelp. When finetuning, we perform an additional 10K optimization steps using the Yelp dataset.  Note that all the baselines require training on in-domain data, while this is optional for our model.  In Figure~\ref{fig:generation}
we find that the \textsc{Autobot} model not exposed to the Yelp dataset during finetuning performed on par with the 
DAE that was trained specifically on Yelp. Additionally, \textsc{Autobot} outperforms the DAE in the above-40 percent accuracy range when finetuned on in-domain data.
We include \textsc{autobot} results with partial finetuning of the encoder in the appendix, which we find considerably improves the Self-BLEU metric.
Since \textsc{AUTOBOT} uses vector arithmetic, inference is as fast as the DAE and over twice that of other methods.

\vspace{-3mm}
\section{Related Work}
Reconstructing text with autoencoders is an active area of research that has lead to several advancements
such as denoising~\cite{vincent2010stacked}, variational~\cite{kingma2013autoencoding, Higgins2017betaVAELB, dai2019diagnosing}, adversarial~\cite{makhzani2015adversarial, zhao2018adversarially}, and regularized~\cite{Ghosh2020From} autoencoders. They have been found especially useful in controlled text generation \citep{hu2017toward, logeswaran2018content, bowman2016generating}, especially in sentiment style transfer \citep{mai2020plug, shen2017style}.

The encoder-decoder structure for obtaining representations has been used in pretraining \citep{lewis2019bart}, sentence infilling \citep{huang2020inset}, and multilingual \citep{artetxe18} scenarios. In particular, \citet{lewis2019bart} treat denoising as translation task to perform pretraining from scratch, but their approach does not induce a sentence representation space with generative properties.
In contrast, our method makes use of a frozen pretrained transformer to learn a shallow, sentence bottleneck autoencoder on top.





\section{Conclusion}
We proposed an approach that converts a pretrained transformer language model into a  
sentence-level autoencoder that is able to reconstruct its pretraining data. The resulting model improves the performance of the pretrained model on sentence-level tasks while maintaining its performance on multi-sentence tasks. In addition, the new sentence representations are suitable for efficient conditional text generation such as sentiment transfer without the need for training on in-domain data. 


\section*{Acknowledgments}
The authors thank Jungo Kasai 
and the anonymous reviewers for their helpful feedback.
Nikolaos Pappas was supported by the Swiss National Science Foundation grant P400P2\_183911. 

\bibliography{main}

\begin{thebibliography}{38}
\expandafter\ifx\csname natexlab\endcsname\relax\def\natexlab#1{#1}\fi

\bibitem[{Artetxe and Schwenk(2019)}]{artetxe18}
Mikel Artetxe and Holger Schwenk. 2019.
\newblock Massively multilingual sentence embeddings for zero-shot
  cross-lingual transfer and beyond.
\newblock \emph{Transactions of the Association for Computational Linguistics},
  7:597--610.

\bibitem[{Bowman et~al.(2016{\natexlab{a}})Bowman, Vilnis, Vinyals, Dai,
  Jozefowicz, and Bengio}]{bowman2016generating}
Samuel Bowman, Luke Vilnis, Oriol Vinyals, Andrew Dai, Rafal Jozefowicz, and
  Samy Bengio. 2016{\natexlab{a}}.
\newblock Generating sentences from a continuous space.
\newblock In \emph{Proceedings of The 20th SIGNLL Conference on Computational
  Natural Language Learning}, pages 10--21.

\bibitem[{Bowman et~al.(2015)Bowman, Angeli, Potts, and
  Manning}]{bowman2015snli}
Samuel~R. Bowman, Gabor Angeli, Christopher Potts, and Christopher~D. Manning.
  2015.
\newblock A large annotated corpus for learning natural language inference.
\newblock In \emph{Proceedings of the 2015 Conference on Empirical Methods in
  Natural Language Processing (EMNLP)}. Association for Computational
  Linguistics.

\bibitem[{Bowman et~al.(2016{\natexlab{b}})Bowman, Vilnis, Vinyals, Dai,
  Jozefowicz, and Bengio}]{bowman2015generating}
Samuel~R. Bowman, Luke Vilnis, Oriol Vinyals, Andrew Dai, Rafal Jozefowicz, and
  Samy Bengio. 2016{\natexlab{b}}.
\newblock Generating sentences from a continuous space.
\newblock In \emph{Proceedings of The 20th {SIGNLL} Conference on Computational
  Natural Language Learning}, pages 10--21, Berlin, Germany. Association for
  Computational Linguistics.

\bibitem[{Brown et~al.(2020)Brown, Mann, Ryder, Subbiah, Kaplan, Dhariwal,
  Neelakantan, Shyam, Sastry, Askell et~al.}]{brown2020language}
Tom~B Brown, Benjamin Mann, Nick Ryder, Melanie Subbiah, Jared Kaplan, Prafulla
  Dhariwal, Arvind Neelakantan, Pranav Shyam, Girish Sastry, Amanda Askell,
  et~al. 2020.
\newblock Language models are few-shot learners.
\newblock \emph{arXiv preprint arXiv:2005.14165}.

\bibitem[{Cer et~al.(2017)Cer, Diab, Agirre, Lopez-Gazpio, and
  Specia}]{cer2017semeval}
Daniel Cer, Mona Diab, Eneko Agirre, I{\~n}igo Lopez-Gazpio, and Lucia Specia.
  2017.
\newblock {S}em{E}val-2017 task 1: Semantic textual similarity multilingual and
  crosslingual focused evaluation.
\newblock In \emph{Proceedings of the 11th International Workshop on Semantic
  Evaluation ({S}em{E}val-2017)}, pages 1--14, Vancouver, Canada. Association
  for Computational Linguistics.

\bibitem[{Conneau et~al.(2020)Conneau, Khandelwal, Goyal, Chaudhary, Wenzek,
  Guzm{\'a}n, Grave, Ott, Zettlemoyer, and Stoyanov}]{conneau2020unsupervised}
Alexis Conneau, Kartikay Khandelwal, Naman Goyal, Vishrav Chaudhary, Guillaume
  Wenzek, Francisco Guzm{\'a}n, {\'E}douard Grave, Myle Ott, Luke Zettlemoyer,
  and Veselin Stoyanov. 2020.
\newblock Unsupervised cross-lingual representation learning at scale.
\newblock In \emph{Proceedings of the 58th Annual Meeting of the Association
  for Computational Linguistics}, pages 8440--8451.

\bibitem[{Conneau et~al.(2017)Conneau, Kiela, Schwenk, Barrault, and
  Bordes}]{conneau2017supervised}
Alexis Conneau, Douwe Kiela, Holger Schwenk, Lo{\"\i}c Barrault, and Antoine
  Bordes. 2017.
\newblock Supervised learning of universal sentence representations from
  natural language inference data.
\newblock In \emph{EMNLP}.

\bibitem[{Dai and Wipf(2019)}]{dai2019diagnosing}
Bin Dai and David Wipf. 2019.
\newblock Diagnosing and enhancing {VAE} models.
\newblock In \emph{International Conference on Learning Representations}.

\bibitem[{Dathathri et~al.(2020)Dathathri, Madotto, Lan, Hung, Frank, Molino,
  Yosinski, and Liu}]{Dathathri2020Plug}
Sumanth Dathathri, Andrea Madotto, Janice Lan, Jane Hung, Eric Frank, Piero
  Molino, Jason Yosinski, and Rosanne Liu. 2020.
\newblock Plug and play language models: A simple approach to controlled text
  generation.
\newblock In \emph{International Conference on Learning Representations}.

\bibitem[{Devlin et~al.(2019)Devlin, Chang, Lee, and
  Toutanova}]{devlin-etal-2019-bert}
Jacob Devlin, Ming-Wei Chang, Kenton Lee, and Kristina Toutanova. 2019.
\newblock {BERT}: Pre-training of deep bidirectional transformers for language
  understanding.
\newblock In \emph{Proceedings of the 2019 Conference of the North {A}merican
  Chapter of the Association for Computational Linguistics: Human Language
  Technologies, Volume 1 (Long and Short Papers)}, pages 4171--4186,
  Minneapolis, Minnesota. Association for Computational Linguistics.

\bibitem[{Ghosh et~al.(2020)Ghosh, Sajjadi, Vergari, Black, and
  Scholkopf}]{Ghosh2020From}
Partha Ghosh, Mehdi S.~M. Sajjadi, Antonio Vergari, Michael Black, and Bernhard
  Scholkopf. 2020.
\newblock From variational to deterministic autoencoders.
\newblock In \emph{International Conference on Learning Representations}.

\bibitem[{Higgins et~al.(2017)Higgins, Matthey, Pal, Burgess, Glorot,
  Botvinick, Mohamed, and Lerchner}]{Higgins2017betaVAELB}
Irina Higgins, Lo{\"i}c Matthey, Arka Pal, Christopher Burgess, Xavier Glorot,
  Matthew~M Botvinick, Shakir Mohamed, and Alexander Lerchner. 2017.
\newblock beta-vae: Learning basic visual concepts with a constrained
  variational framework.
\newblock In \emph{International Conference on Learning Representations}.

\bibitem[{Hochreiter and Schmidhuber(1997)}]{hochreiter1997long}
Sepp Hochreiter and J{\"u}rgen Schmidhuber. 1997.
\newblock Long short-term memory.
\newblock \emph{Neural computation}, 9(8):1735--1780.

\bibitem[{Hu et~al.(2017)Hu, Yang, Liang, Salakhutdinov, and
  Xing}]{hu2017toward}
Zhiting Hu, Zichao Yang, Xiaodan Liang, Ruslan Salakhutdinov, and Eric~P Xing.
  2017.
\newblock Toward controlled generation of text.
\newblock In \emph{ICML}.

\bibitem[{Huang et~al.(2020)Huang, Zhang, Elachqar, and Cheng}]{huang2020inset}
Yichen Huang, Yizhe Zhang, Oussama Elachqar, and Yu~Cheng. 2020.
\newblock Inset: Sentence infilling with inter-sentential transformer.
\newblock In \emph{Proceedings of the 58th Annual Meeting of the Association
  for Computational Linguistics}, pages 2502--2515.

\bibitem[{Kasai et~al.(2021)Kasai, Pappas, Peng, Cross, and
  Smith}]{kasai2021deep}
Jungo Kasai, Nikolaos Pappas, Hao Peng, James Cross, and Noah Smith. 2021.
\newblock Deep encoder, shallow decoder: Reevaluating non-autoregressive
  machine translation.
\newblock In \emph{International Conference on Learning Representations}.

\bibitem[{Kingma and Welling(2014)}]{kingma2013autoencoding}
Diederik~P Kingma and Max Welling. 2014.
\newblock Auto-encoding variational bayes.
\newblock In \emph{International Conference on Learning Representations}.

\bibitem[{Krause et~al.(2021)Krause, Gotmare, McCann, Keskar, Joty, richard
  socher, and Rajani}]{krause2021gedi}
Ben Krause, Akhilesh~Deepak Gotmare, Bryan McCann, Nitish~Shirish Keskar,
  Shafiq Joty, richard socher, and Nazneen Rajani. 2021.
\newblock Gedi: Generative discriminator guided sequence generation.

\bibitem[{Lewis et~al.(2019)Lewis, Liu, Goyal, Ghazvininejad, Mohamed, Levy,
  Stoyanov, and Zettlemoyer}]{lewis2019bart}
Mike Lewis, Yinhan Liu, Naman Goyal, Marjan Ghazvininejad, Abdelrahman Mohamed,
  Omer Levy, Ves Stoyanov, and Luke Zettlemoyer. 2019.
\newblock Bart: Denoising sequence-to-sequence pre-training for natural
  language generation, translation, and comprehension.
\newblock \emph{arXiv preprint arXiv:1910.13461}.

\bibitem[{Li et~al.(2020)Li, Zhou, He, Wang, Yang, and
  Li}]{li-etal-2020-sentence}
Bohan Li, Hao Zhou, Junxian He, Mingxuan Wang, Yiming Yang, and Lei Li. 2020.
\newblock On the sentence embeddings from pre-trained language models.
\newblock In \emph{Proceedings of the 2020 Conference on Empirical Methods in
  Natural Language Processing (EMNLP)}, pages 9119--9130, Online. Association
  for Computational Linguistics.

\bibitem[{Liu et~al.(2019)Liu, Ott, Goyal, Du, Joshi, Chen, Levy, Lewis,
  Zettlemoyer, and Stoyanov}]{liu2019RoBERTa}
Y.~Liu, Myle Ott, Naman Goyal, Jingfei Du, Mandar Joshi, Danqi Chen, Omer Levy,
  M.~Lewis, Luke Zettlemoyer, and Veselin Stoyanov. 2019.
\newblock Roberta: A robustly optimized bert pretraining approach.
\newblock \emph{ArXiv}, abs/1907.11692.

\bibitem[{Logeswaran et~al.(2018)Logeswaran, Lee, and
  Bengio}]{logeswaran2018content}
Lajanugen Logeswaran, Honglak Lee, and Samy Bengio. 2018.
\newblock Content preserving text generation with attribute controls.
\newblock \emph{Advances in Neural Information Processing Systems}, 31.

\bibitem[{Mai et~al.(2020)Mai, Pappas, Montero, Smith, and
  Henderson}]{mai2020plug}
Florian Mai, Nikolaos Pappas, Ivan Montero, Noah~A Smith, and James Henderson.
  2020.
\newblock Plug and play autoencoders for conditional text generation.
\newblock In \emph{Proceedings of the 2020 Conference on Empirical Methods in
  Natural Language Processing (EMNLP)}, pages 6076--6092.

\bibitem[{Makhzani et~al.(2016)Makhzani, Shlens, Jaitly, and
  Goodfellow}]{makhzani2015adversarial}
Alireza Makhzani, Jonathon Shlens, Navdeep Jaitly, and Ian Goodfellow. 2016.
\newblock Adversarial autoencoders.
\newblock In \emph{International Conference on Learning Representations}.

\bibitem[{Peters et~al.(2018)Peters, Neumann, Iyyer, Gardner, Clark, Lee, and
  Zettlemoyer}]{peters-etal-2018-deep}
Matthew Peters, Mark Neumann, Mohit Iyyer, Matt Gardner, Christopher Clark,
  Kenton Lee, and Luke Zettlemoyer. 2018.
\newblock Deep contextualized word representations.
\newblock In \emph{Proceedings of the 2018 Conference of the North {A}merican
  Chapter of the Association for Computational Linguistics: Human Language
  Technologies, Volume 1 (Long Papers)}, pages 2227--2237, New Orleans,
  Louisiana. Association for Computational Linguistics.

\bibitem[{Radford et~al.(2019)Radford, Wu, Child, Luan, Amodei, and
  Sutskever}]{radford2019language}
Alec Radford, Jeffrey Wu, Rewon Child, David Luan, Dario Amodei, and Ilya
  Sutskever. 2019.
\newblock Language models are unsupervised multitask learners.
\newblock \emph{OpenAI blog}, 1(8):9.

\bibitem[{Reimers and Gurevych(2019)}]{Reimers2019SentenceBERT}
Nils Reimers and Iryna Gurevych. 2019.
\newblock Sentence-bert: Sentence embeddings using siamese bert-networks.
\newblock In \emph{EMNLP/IJCNLP}.

\bibitem[{Shen et~al.(2017)Shen, Lei, Barzilay, and Jaakkola}]{shen2017style}
Tianxiao Shen, Tao Lei, Regina Barzilay, and Tommi Jaakkola. 2017.
\newblock Style transfer from non-parallel text by cross-alignment.
\newblock In \emph{Proceedings of the 31st International Conference on Neural
  Information Processing Systems}, pages 6833--6844.

\bibitem[{Shen et~al.(2020{\natexlab{a}})Shen, Mueller, Barzilay, and
  Jaakkola}]{pmlr-v119-shen20c}
Tianxiao Shen, Jonas Mueller, Dr.Regina Barzilay, and Tommi Jaakkola.
  2020{\natexlab{a}}.
\newblock Educating text autoencoders: Latent representation guidance via
  denoising.
\newblock In \emph{Proceedings of the 37th International Conference on Machine
  Learning}, volume 119 of \emph{Proceedings of Machine Learning Research},
  pages 8719--8729. PMLR.

\bibitem[{Shen et~al.(2020{\natexlab{b}})Shen, Mueller, Barzilay, and
  Jaakkola}]{shen2019educating}
Tianxiao Shen, Jonas Mueller, Regina Barzilay, and Tommi Jaakkola.
  2020{\natexlab{b}}.
\newblock Educating text autoencoders: Latent representation guidance via
  denoising.
\newblock In \emph{International Conference on Machine Learning}, pages
  8719--8729. PMLR.

\bibitem[{Song et~al.(2019)Song, Tan, Qin, Lu, and Liu}]{pmlr-v97-song19d}
Kaitao Song, Xu~Tan, Tao Qin, Jianfeng Lu, and Tie-Yan Liu. 2019.
\newblock {MASS}: Masked sequence to sequence pre-training for language
  generation.
\newblock In \emph{Proceedings of the 36th International Conference on Machine
  Learning}, volume~97 of \emph{Proceedings of Machine Learning Research},
  pages 5926--5936. PMLR.

\bibitem[{Vaswani et~al.(2017)Vaswani, Shazeer, Parmar, Uszkoreit, Jones,
  Gomez, Kaiser, and Polosukhin}]{vaswani17}
Ashish Vaswani, Noam Shazeer, Niki Parmar, Jakob Uszkoreit, Llion Jones,
  Aidan~N Gomez, \L~ukasz Kaiser, and Illia Polosukhin. 2017.
\newblock Attention is all you need.
\newblock In I.~Guyon, U.~V. Luxburg, S.~Bengio, H.~Wallach, R.~Fergus,
  S.~Vishwanathan, and R.~Garnett, editors, \emph{Advances in Neural
  Information Processing Systems 30}, pages 5998--6008. Curran Associates, Inc.

\bibitem[{Vincent et~al.(2010)Vincent, Larochelle, Lajoie, Bengio, Manzagol,
  and Bottou}]{vincent2010stacked}
Pascal Vincent, Hugo Larochelle, Isabelle Lajoie, Yoshua Bengio, Pierre-Antoine
  Manzagol, and L{\'e}on Bottou. 2010.
\newblock Stacked denoising autoencoders: Learning useful representations in a
  deep network with a local denoising criterion.
\newblock \emph{Journal of machine learning research}, 11(12).

\bibitem[{Wang et~al.(2018)Wang, Singh, Michael, Hill, Levy, and
  Bowman}]{wang2018glue}
Alex Wang, Amanpreet Singh, Julian Michael, Felix Hill, Omer Levy, and Samuel
  Bowman. 2018.
\newblock Glue: A multi-task benchmark and analysis platform for natural
  language understanding.
\newblock In \emph{Proceedings of the 2018 EMNLP Workshop BlackboxNLP:
  Analyzing and Interpreting Neural Networks for NLP}, pages 353--355.

\bibitem[{Wang et~al.(2019)Wang, Hua, and Wan}]{wang2019controllable}
Ke~Wang, Hang Hua, and Xiaojun Wan. 2019.
\newblock Controllable unsupervised text attribute transfer via editing
  entangled latent representation.
\newblock In \emph{Advances in Neural Information Processing Systems}, pages
  11034--11044.

\bibitem[{Zhao et~al.(2018)Zhao, Kim, Zhang, Rush, LeCun
  et~al.}]{zhao2018adversarially}
Junbo~Jake Zhao, Yoon Kim, Kelly Zhang, Alexander~M Rush, Yann LeCun, et~al.
  2018.
\newblock Adversarially regularized autoencoders.
\newblock In \emph{Proceedings of the 35th International Conference on Machine
  Learning}, Proceedings of Machine Learning Research. PMLR.

\bibitem[{Zhu et~al.(2015)Zhu, Kiros, Zemel, Salakhutdinov, Urtasun, Torralba,
  and Fidler}]{zhu2015aligning}
Yukun Zhu, Ryan Kiros, Rich Zemel, Ruslan Salakhutdinov, Raquel Urtasun,
  Antonio Torralba, and Sanja Fidler. 2015.
\newblock Aligning books and movies: Towards story-like visual explanations by
  watching movies and reading books.
\newblock In \emph{Proceedings of the IEEE international conference on computer
  vision}, pages 19--27.

\end{thebibliography}
\bibliographystyle{acl_natbib}

\clearpage
\appendix
\label{sec:appendix}

\section{Reproducibility}

\subsection{Experimental Setup}

\paragraph{Computing Infrastructure}
For all of our experiments involving base models, we use a computation cluster with 5 NVIDIA RTX 2080 TI GPU, 11GB GPU memory, and 128GB RAM. For large models, we use a computation cluster with 4 NVIDIA TITAN RTX GPUs, 24GB GPU memory and 256GB RAM.

\paragraph{Implementation}
We will make our implementation available on Github.\footnote{\url{https://github.com/ivanmontero/autobot}} We used Python 3.7, PyTorch  1.6.0, and Sentence Transformers 0.3.7. We use modified versions of Fairseq 0.9.0 and Transformers 3.3.1. We obtain our datasets from the citations specified in the main paper.

\paragraph{\textsc{Autobot} Training}\label{adx:autobottraining}
We extract the sentences from the BooksCorpus and English Wikipedia datasets to recreate the BERT dataset, and use RoBERTa-base's pretrained tokenizer for tokenization. We report our hyperparameters for \textsc{Autobot}-base in Table~\ref{tab:autobotbase}. Our decoder only has one single layer, and RoBERTa-base remains fixed during finetuning.

\begin{table}[h]
    \small
    \centering
    \begin{tabular}{ll}
        \toprule
        \textsc{Model Parameters} & \textsc{Value} \\
        \midrule
        \textbf{Fixed Parameters} & {} \\
        \midrule
        Transformer Encoder & RoBERTa-base \\
        Transformer Encoder Fixed & True \\
        Warmup Steps & 4000 \\
        Dropout & 0.1 \\
        Optimizer & Adam \\
        Learning Rate Schedule & Linear Decay \\
        Max Sequence Length & 128 \\
        Max Tokens & 24576 \\
        Bottleneck Heads & 12 \\
        Hidden Size & 768 \\
        Decoder Layers & 1 \\
        \midrule
        \textbf{Tuned Parameters} & {} \\
        \midrule
        Num Steps & $\{1k, 10k, 100k\}$ \\
        Learning Rate & $\{$1e-3, 1e-4, 1e-5$\}$ \\
        \midrule
        \textbf{Optimal Parameters} & {} \\
        \midrule
        Num Steps & $100k$ \\
        Learning Rate & $1e-3$ \\
        \midrule
        \textbf{Extra Info} & {} \\
        \midrule
        Model Size (\# params) & $127M$ \\
        \bottomrule
    \end{tabular}
    \caption{\label{tab:autobotbase}Hyperparamters for \textsc{AUTOBOT}-base}

\end{table}

\subsection{Sentence Representations}\label{adx:sentrep}

We use the Sentence Transformers framework for training and evaluation of \textsc{Autobot}. We use the default settings in their framework to train on NLI, and evaluate using the Spearman correlation of the cosine similarity. During NLI finetuning, we only use the encoder and bottlneck, with the bottleneck representation used as the sentence representation, and allow for all parameters to be finetuned.

\subsection{Sentence Generation}

We use a modified version of Fairseq's generation code for encoder-decoder models to perform vector arithmetic for sentiment transfer. We follow the instructions of \citet{mai2020plug} to finetune a sentiment classifier using DistilBERT from the Huggingface transformers library.

For the \textsc{Autobot} models finetuned to the Yelp dataset, we follow the exact same steps as Appendix~\ref{adx:autobottraining} except beginning with the \textsc{Autobot}-base model, using the Yelp training set, and performing 10k optimization steps.

\subsection{Sentence Classification}

We use the Huggingface library to perform sentence classification using \textsc{Autobot}.  During finetuning, we only use the encoder and bottleneck, with the bottleneck representation used as a \texttt{CLS} representation, and allow for all parameters to be finetuned. We perform a hyperparameter search similar to that of RoBERTa by comparing development performances when using $\{$1e-5, 2e-5, 3e-5$\}$ for the learning rate.


\section{Additional Results}
We provide additional results in addition to our experiments below.

\subsection{Autoencoding Steps}

We perform an ablation study on the effect of autoencoding finetuning steps of the underlying pretrained encoder during autoencoding on the downstream sentence representation performance. We provide the detailed performances of performing Table~\ref{tab:autobotbase} when using a learning rate of 1e-3 in Table~\ref{tab:autosteps}.

\begin{table}[ht!]
	\centering 
	\footnotesize
	\begin{tabular}{ l | c }
		\toprule
		\textbf{Training Steps} & \textbf{Spearman} \\ \midrule
		1 & 74.38 \\
		1k & 75.45 \\
		10k & 78.01 \\
		100k & \textbf{78.59}  \\
		\hline
		baseline & 77.03 \\
        \bottomrule
	\end{tabular}
	\caption{\label{tab:autosteps}\textsc{Autobot} pretraining steps vs. sentence representation performance when training on NLI and evaluating on STS}
	
\end{table}

\subsection{Finetunable Encoder Layers}\label{adx:finetune}
We perform an ablation study on the effect of finetuning the underlying pretrained encoder during autoencoding on downstream sentence representation performance. We provide the detailed performances of performing Table~\ref{tab:autobotbase} with the optimal parameters, but varying how many of the last layers of RoBERTa-base to finetune. Results are in Table~\ref{tab:finetune}

\begin{table}[ht!]
	\centering 
	\footnotesize

	\begin{tabular}{ l | c }
		\toprule
		\textbf{Finetunable Layers} & \textbf{Spearman} \\ \midrule
		None & \textbf{78.59} \\
		1 & 77.24 \\
		2 & 76.17 \\
		3 & 76.20 \\
		\hline
		baseline & 77.03 \\
        \bottomrule
	\end{tabular}
	\caption{\label{tab:finetune} \textsc{Autobot} finetunable layers vs. sentence representation performance when training on NLI and evaluating on STS}
	
\end{table}

\subsection{Finetunable Encoder Generation}
We provide an appended generation table from Section~\ref{sec:gen} to include the generation results we obtained by allowing the top three layers of RoBERTa-base to be finetuned during autoencoding on the style generation task. The results are shown in Figure~\ref{fig:gen_plus_finetune}. The same model as used in Appendix~\ref{adx:finetune} is used.
\begin{figure}[ht]
\centering
\hspace{-2mm}\includegraphics[width=0.49\textwidth]{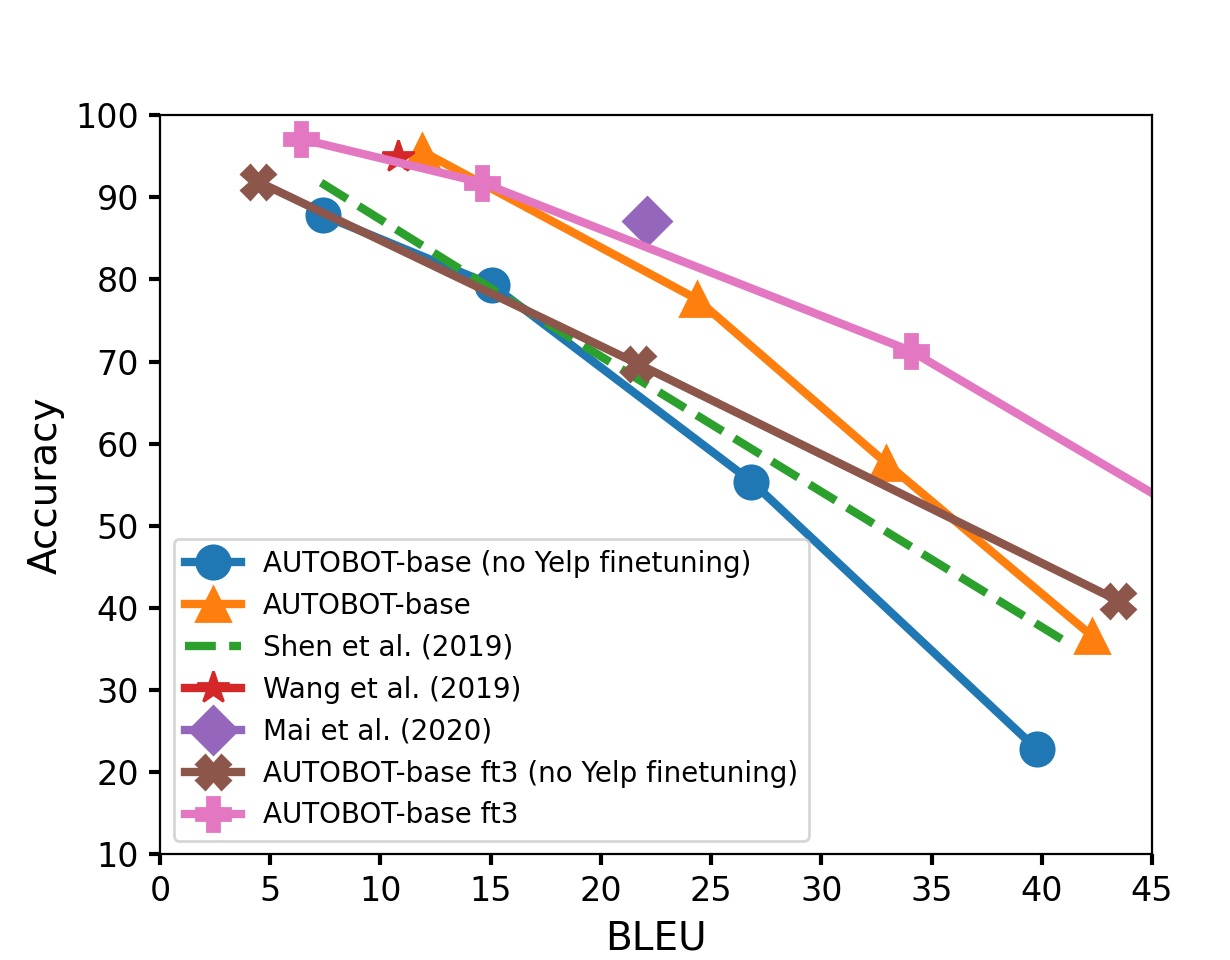} 
\caption{\label{fig:gen_plus_finetune} 
Automatic evaluations of vector arithmetic for sentiment transfer, plotted as accuracy vs. self-BLEU. Accuracy is measured by a sentiment classifier, and values for varying multiples of the sentiment vector are plotted. Upper right is better.
}
\end{figure}

\subsection{Style Transfer Results}
We provide Table~\ref{tab:pnp} that reports results on the Yelp sentiment transfer test set from the generation table in Section~\ref{sec:gen}, appending to the table \citep{mai2020plug}. We outline the relative time differences during inference. We can observe that our model not only provides competitive speed-quality tradeoff.

\begin{table}[ht!]
    \centering
    \footnotesize
  \def\arraystretch{1.1}\tabcolsep=3.5pt
    \begin{tabular}{l|r|r|r}
        \toprule
        \textbf{Model} & \textbf{Acc.} & \textbf{BLEU} & \textbf{+Time} \\
        \toprule
        \hline
        FGIM & 94.9 & 10.8 & 70.0$\times$ \\
        Mai et al. 2020 + FGIM & 93.1 & 18.1 & 2820.0$\times$ \\ 
        Mai et al. 2020 & 87.1 & 22.1  & 1.0$\times$ \\ \hline
        Shen et al. (2019)&  96.8 & 6.5 & 0.5$\times$ \\ 
        AUTOBOT-base (ours) & 95.6 & 11.90 & 0.5$\times$\\\bottomrule
    \end{tabular}\vspace{-1mm}
    \caption{Self BLEU on the Yelp sentiment transfer test set
    with highest transfer accuracy (``Acc.'').  
    ``+Time'' reports the inference-time slowdown factor due to each method's additional computation relative to the method by \citet{mai2020plug}.
    }\vspace{-3mm}
    \label{tab:pnp}
\end{table}  

\subsection{Detailed Sentence Classification Results}
Section~\ref{sec:clf} provides a summary of the GLUE results, while outlining the specific single-sentence classification performances. We provide the results for each task in Table~\ref{tab:roberta_glue}


\begin{table*}[ht]
\centering
\begin{tabular}{lccccccccc}
\toprule
& \bf MNLI & \bf QNLI & \bf QQP & \bf RTE & \bf SST & \bf MRPC & \bf CoLA & \bf STS \\
\midrule 
$\text{RoBERTa-base}$ & 87.6 & 92.8 & 91.9 & 78.7 & 94.8 & 90.2 & 63.6 & 91.2 \\
$\text{AUTOBOT-base}$ & 88.0 & 92.7 & 91.9 & 79.5 & 95.0 & 88.4 & 66.0 & 91.4 \\
\midrule 
$\text{RoBERTa-large}$ & 90.2 & 94.7 & 92.2 & 86.6 & 96.4 & 90.9 & 68.0 & 92.4 \\
$\text{AUTOBOT-large}$ & 90.5 & 94.6 & 92.2 & 87.6 & 96.9 & 89.0 & 70.2 & 92.4 \\
\bottomrule
\end{tabular} 
\caption{
Dev.~results on GLUE.
For RTE, STS and MRPC we finetune starting from the MNLI model instead of the baseline pretrained model.}

\label{tab:roberta_glue}
\end{table*}

\end{document}